\documentclass[conference]{IEEEtran}
\IEEEoverridecommandlockouts
\usepackage{cite}
\usepackage{amsmath,amssymb,amsfonts}
\usepackage{algorithmic}
\usepackage{graphicx}
\usepackage{textcomp}

\newcommand{\re}[1]{{\color{black}#1}}
\usepackage{xcolor}
\def\BibTeX{{\rm B\kern-.05em{\sc i\kern-.025em b}\kern-.08em
    T\kern-.1667em\lower.7ex\hbox{E}\kern-.125emX}}
\begin{document}

\title{Transformer-based Drum-level Prediction  
in a Boiler Plant with Delayed Relations among Multivariates\\

}

\author{\IEEEauthorblockN{Gang Su}
\IEEEauthorblockA{\textit{SenseTime Research} \\
China \\
sugang@sensetime.com}
\and
\IEEEauthorblockN{Sun Yang}
\IEEEauthorblockA{\textit{School of Software Engineering} \\
\textit{Peking University}\\
Beijing, China \\
yangsun@sensetime.com}
\and
\IEEEauthorblockN{Zhishuai Li}
\IEEEauthorblockA{\textit{SenseTime Research} \\
China \\
lizhishuai@sensetime.com}
\and
\IEEEauthorblockN{Ziyue Li $^\dagger$}
\IEEEauthorblockA{\textit{Information Systems} \\
\textit{University of Cologne}\\
Cologne, Germany \\
zlibn@wiso.uni-koeln.de}
\thanks{$^\dagger$: corresponding author}
}

\maketitle

\begin{abstract}
The steam drum water level is a critical parameter that directly impacts the safety and efficiency of power plant operations. However, predicting the drum water level in boilers is challenging due to complex non-linear process dynamics \re{originating from long-time delays and interrelations}, as well as measurement noise. This paper investigates the application of Transformer-based models for predicting drum water levels in a steam boiler plant. Leveraging the capabilities of Transformer architectures, this study aims to develop an accurate and robust predictive framework to anticipate water level fluctuations and facilitate proactive control strategies. \re{To this end, a prudent pipeline is proposed, including 1) data preprocess, 2) causal relation analysis, 3) delay inference, 4) variable augmentation, and 5) prediction}. Through extensive experimentation and analysis, the effectiveness of Transformer-based approaches in steam drum water level prediction is evaluated, highlighting their potential to enhance operational stability and optimize plant performance.
\end{abstract}



\section{Introduction}

The efficient operation of steam boilers in power plants relies heavily on maintaining optimal water levels within the boiler drum. Proper control of drum level is essential for ensuring continuous steam production while adhering to safety regulations. However, achieving precise control of drum level is a challenging task due to various process disturbances and dynamics inherent in boiler systems. Traditionally, control strategies for drum level regulation have relied on feedback and feedforward control techniques, often employing Proportional Integral Derivative (PID \cite{ang2005pid}) controllers in conjunction with rule-based feedforward controllers. While these strategies have been effective to some extent, they often struggle to adapt to changing operating conditions and fail to capture the complex dynamics of the boiler system. 

In recent years, there has been a growing interest in leveraging advanced deep-learning techniques to predict the future drum level, \re{which enhances the PID via feedforward compensation control.} 

\begin{figure}[t]
\centerline{\includegraphics[width=\linewidth]{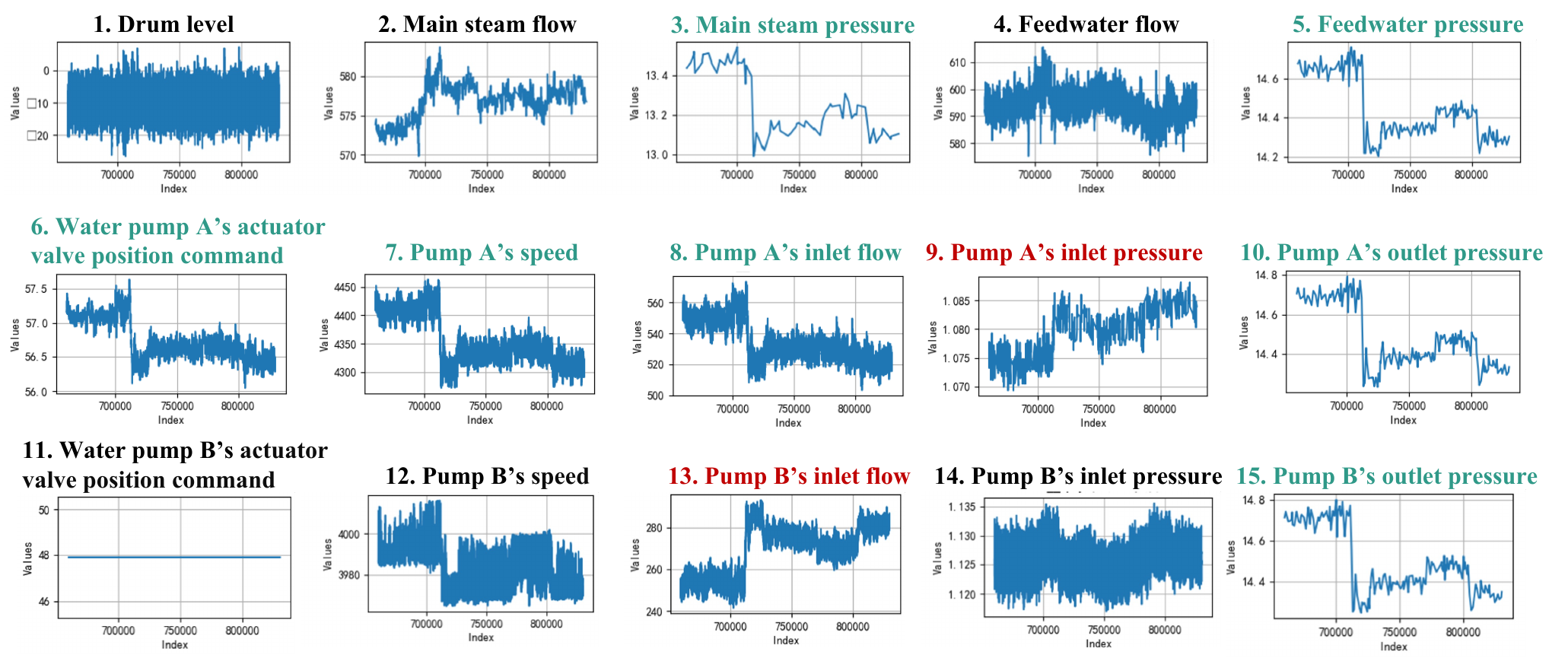}}
\caption{Visualization of selected variables.}
\label{fig:variables}
\vspace{-10pt}
\end{figure}

\re{However, to accurately predict the drum level is rather challenging given the multiple physico-chemical factors are intercorrelated, with complex and delayed interrelations. As visualized in Fig. \ref{fig:variables}, 15 out of in total 62 variables are selected for visualization, we can clearly observe that: (1) some variables are following the same ``stepping-down'' pattern, i.e., the variables in green font, e.g., variables 3, 5, 6, etc., (2) some variables are in an opposite ``stepping-up'' trend, i.e., the variables in red font, e.g., variables 9, 13, etc. There is usually physical information behind the complex interrelations: for example, the water pump A's command (variable 6) tunes the A's inlet flow (variable 8) via controlling the pump speed (variable 7), thus they are in the same trend; Pump A and B's outlet are shared via a same parent pipe; thus, their outlet pressures are always in the same trend. To this end, to understand all the variables causal interrelation is critical to predict the final drum level. 

Moreover, given the long reaction chain, for example, increasing pump A's inlet flow will take a long delay to be reflected in the increasing drum level (usually the delay will be around 100 seconds and it varies to different variables and different boiler plant), so when predicting, pump A's inlet flow at time step $t$ should be used to predict the drum level at time step $t + t_{\text{delay}}$. Thus, identifying the delay effect of variables of interest on the drum level is also critical.
}

\re{
In this study, we aim to develop a data-driven model based on Transformer architecture for predicting drum level variations in steam boilers. To tackle the challenges mentioned above, we proposed a delicate framework correspondingly, shown in Fig. \ref{fig:model}, with the key steps of \textbf{causal relation analysis} to screen direct parent factors that affect drum level, and \textbf{delay inference} to find out each factors' delays, and \textbf{variable augmentation} to generate lagged variables according to the delays. Finally, a Transformer model \cite{vaswani2017attention,mao2024pdit,mao2022transformer,jiang2024x,li2023critical} will be used for time-series prediction.}


Through experimentation and analysis, we seek to demonstrate the effectiveness of our proposed approach in real-world boiler systems. By leveraging the capabilities of Transformer-based models and combining them with established control techniques, we aim to revolutionize the way drum level control is implemented in steam boiler plants, paving the way for more adaptive and intelligent control systems in the power generation industry.


\section{Related Work}


In recent years, model predictive control (MPC) has become increasingly popular and widely used in drum boilers \cite{damiran2016model}. MPC is a model-based control method that minimizes a cost function derived from predicted future plant outputs and control actions at each time step. It predicts future system behavior by establishing a system model and combining it with forecasts of future outputs and actions. MPC utilizes current system state and control history to compute expected future behavior, then determines the optimal control sequence to achieve desired system performance. The optimized control input is applied to the plant, with the process repeated at each time step to continuously adjust inputs for dynamic operation. However, MPC's predictive capabilities are typically constrained by its reliance on precise system models, the potential complexity involved in model development, finite prediction horizons, sensitivity to disturbances, and the computational demands of real-time optimization. 

To overcome the limitations of MPC, some researchers \cite{morari2012nonlinear,siddiqui2020offset} introduce an offset-free Nonlinear Model Predictive Controller (NMPC) for the control of a drum-boiler pilot-plant. The NMPC is designed based on a Multi-Input Multi-Output (MIMO) nonlinear model of the pilot plant, developed using first principle equations and thermodynamic properties. An Extended Kalman Filter (EKF) is incorporated for disturbance modeling and state prediction to ensure offset-free control. However, the experiments conducted in the aforementioned methods were based on simulations, overlooking the disparities between simulators and real-world scenarios. Real-world environments often present much greater complexity than simulators. Therefore, this paper will train a prediction model using real plant data.

There has been abundant research in time series prediction, based on various techniques such as statistical models \cite{jiang2023unified}, tensor \cite{li2020tensor,tsung2020discussion,li2020long,li2022profile,sergin2024low,yan2024sparse,ziyue2021tensor,li2022individualized,li2023tensor,li2023choose}, deep learning 
\cite{chen2023adaptive,xu2023kits,guo2024online}, contrastive learning\cite{ruan2023privacy,mao2022jointly,wang2023correlated,li2024non}, and Large Language Models (LLMs) \cite{liu2024spatial,liu2024timecma,zhang2024dualtime}. Several studies have investigated the effectiveness of various machine learning approaches for steam drum water level prediction in power plant boilers, including Neural Network (NN) \cite{oko2015neural,selvi2017modeling}, Recurrent Neural Network (RNN) \cite{Madhavan}, and Long Short-Term Memory (LSTM) \cite{mugweni2021neural}, for the prediction and control of steam drum water levels in power plant boilers. 

However, existing studies may lack the capability to capture complex temporal dependencies inherent in steam drum water level dynamics.  
This paper \re{instead} addresses this gap by \re{proposing a rigid framework to consider the complex interrelations and temporal delays,} utilizing Transformer architectures and causal learning framework \cite{lan2023mm,lin2023dynamic,lan2024multifun} in conjunction with PID controllers in cascade control configurations for improved water level control in power plant boilers.



\section{Methodology}
\begin{figure}[t]
\centerline{\includegraphics[width=\linewidth]{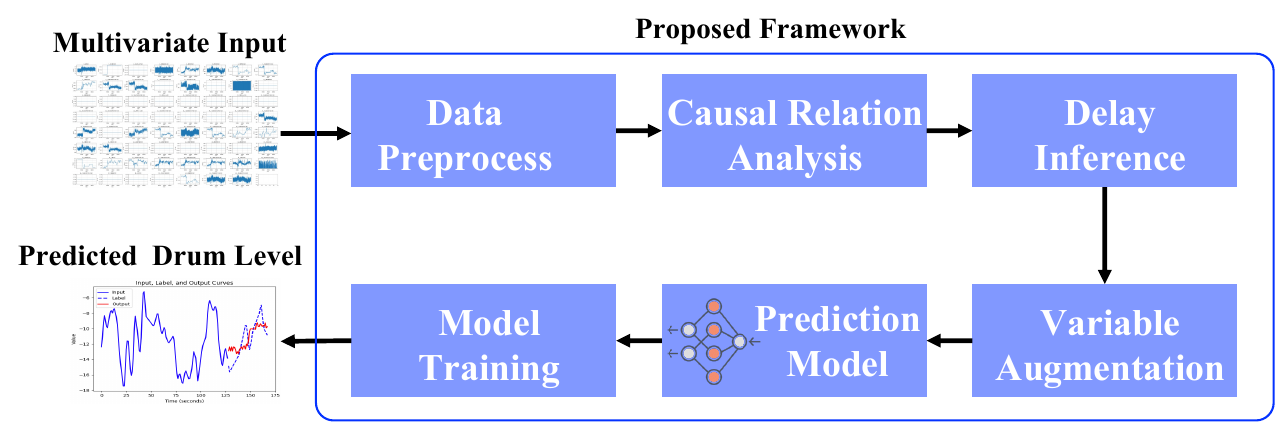}}
\caption{The Proposed Framework.}
\label{fig:model}
\vspace{-10pt}
\end{figure}

\re{
The proposed framework contains the following key steps, as visualized in Fig. \ref{fig:model}:
\begin{enumerate}
    \item The multivariable input will be first preprocessed, including normalization, denoising, and dealing with missing values and anomalous values;
    \item Then, a causal relation analysis based on both domain knowledge and Granger Causality \cite{granger1969investigating} will be conducted. According to the law of conditional independence in causal graphs, specifically Bayesian Network \cite{heckerman2008tutorial}, only the direct parent factors of drum level will be considered for prediction;
    \item The screened factors will be further examined to infer their optimal delay with respect to the drum level;
    \item After inferring the delay for each factor, a few augmentation schemes will be conducted: both the original variable and the one shifted forward by $t_{\text{delay}}$ will be kept for prediction. 
    \item Finally, the processed variables will be fed into the prediction model for training.
\end{enumerate}
}


\subsection{Data Preprocessing and Feature Engineering}\label{AA}
During \textbf{data preprocessing}, the following steps were undertaken:
\begin{itemize}
\item Data Cleaning: Removal of missing values and outliers to ensure data quality and accuracy.
\item Data Smoothing: Smoothing of raw data to reduce noise and fluctuations, enhancing data interpretability.
\item Data Standardization: Standardization of variables to a common scale to eliminate dimensional effects and facilitate model training and convergence. Here, we used LayerNorm \cite{ba2016layer}, a normalization technique that helps maintain stable gradients during the training process. 
 
\end{itemize}

During the \textbf{feature engineering} phase, we focused on identifying key variables that have a significant impact on steam drum water level. These variables include measurements like feedwater flow, feedwater pressure, feedwater temperature, main steam flow, main steam pressure, main steam temperature, coal consumption, generator power, and desuperheating water flow. Leveraging domain knowledge, we then crafted additional variables to enhance the model's predictive capabilities. These new variables capture important relationships and interactions, such as the difference between feedwater flow and main steam flow, the difference between main steam temperature and feedwater temperature, and various pressure differentials. By incorporating these features, we aimed to better capture the complex dynamics of the steam drum system and improve the accuracy of our predictions. Additionally, domain-specific formulas were considered to derive new variables, such as thermal formulas reflecting the heat absorption in the drum.

\begin{figure}[t]
\centering
\includegraphics[width=\columnwidth]{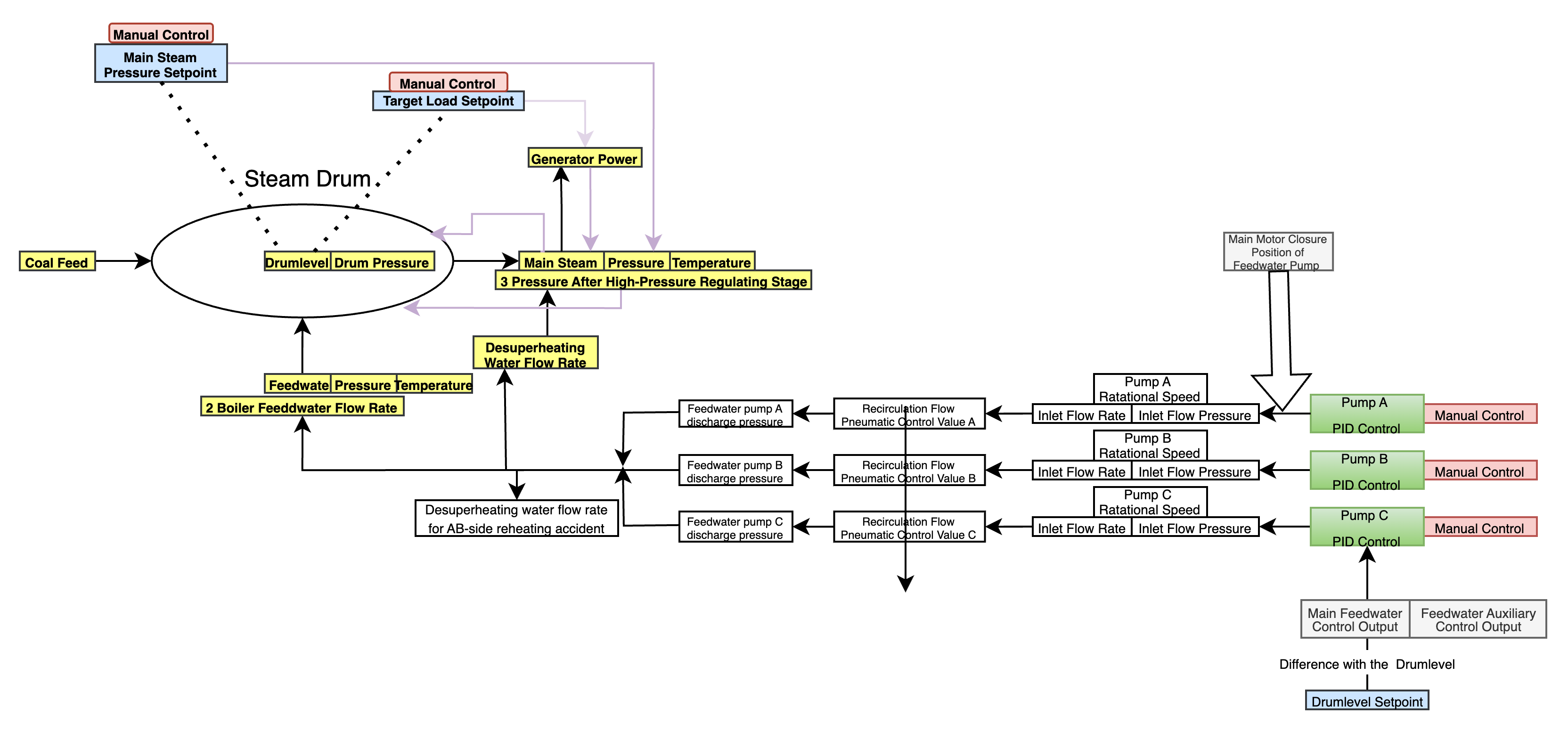}
\caption{A visualization of the steam drum water level control system with relevant variables positioned accordingly: the variables of the most interest are highlighted with color yellow, and the factors in blue and red with a dotted link need to be further analyzed.}
\label{fig:control_system}
\vspace{-10pt}
\end{figure}

\subsection{Causal Relationships}
We will employ Granger causality, assistant with domain knowledge, to identify the factors that have direct causal links pointing to the drum level.

The key idea of Granger causality is: a time series $\mathbf{x}$ is said to be Granger-cause of another time series $\mathbf{y}$ if past values of $\mathbf{x}$ provide significant information about the future values of $\mathbf{y}$ beyond what is already provided by the past values of $\mathbf{y}$ itself. Thus, a statistical test will be conducted by comparing the following two:
\begin{equation}
\begin{split}
    & y_t = f_1(\mathbf{y}_{t-\tau : t-1}, \mathbf{Z}_{t-\tau : t-1}) + e\\
    & y'_t = f_1(\mathbf{y}_{t-\tau : t-1}, \mathbf{Z}_{t-\tau : t-1} + \mathbf{x}_{t-\tau : t-1}) + e'\\
    & x \text{ being the cause of } y = \textit{Wilcoxon-test}(y_t, y'_t)
\end{split}
\end{equation}
where for example, our $y, y'$ is the drum level, $x$ is each factor in Fig. \ref{fig:control_system}, the $\mathbf{Z}$ are the other control variables for prediction, $\tau$ is the history length, $e$ is the error, and $f_1(\cdot)$ is the one-step prediction model, i.e., LSTM or Transformer.

Since LSTM may not follow the assumed F-distribution, the Wilcoxon signed-rank test, a non-parametric test, is used to compare the two 
\cite{swarup2022effects}. 

We tested the factors that have a direct causal link with drum level in Fig. \ref{fig:control_system}, and all their $p$-values are smaller than 0.05, which means that they are statistically significant to be considered as the cause of drum level.

\begin{figure}[htbp]
\centerline{\includegraphics[width=\linewidth]{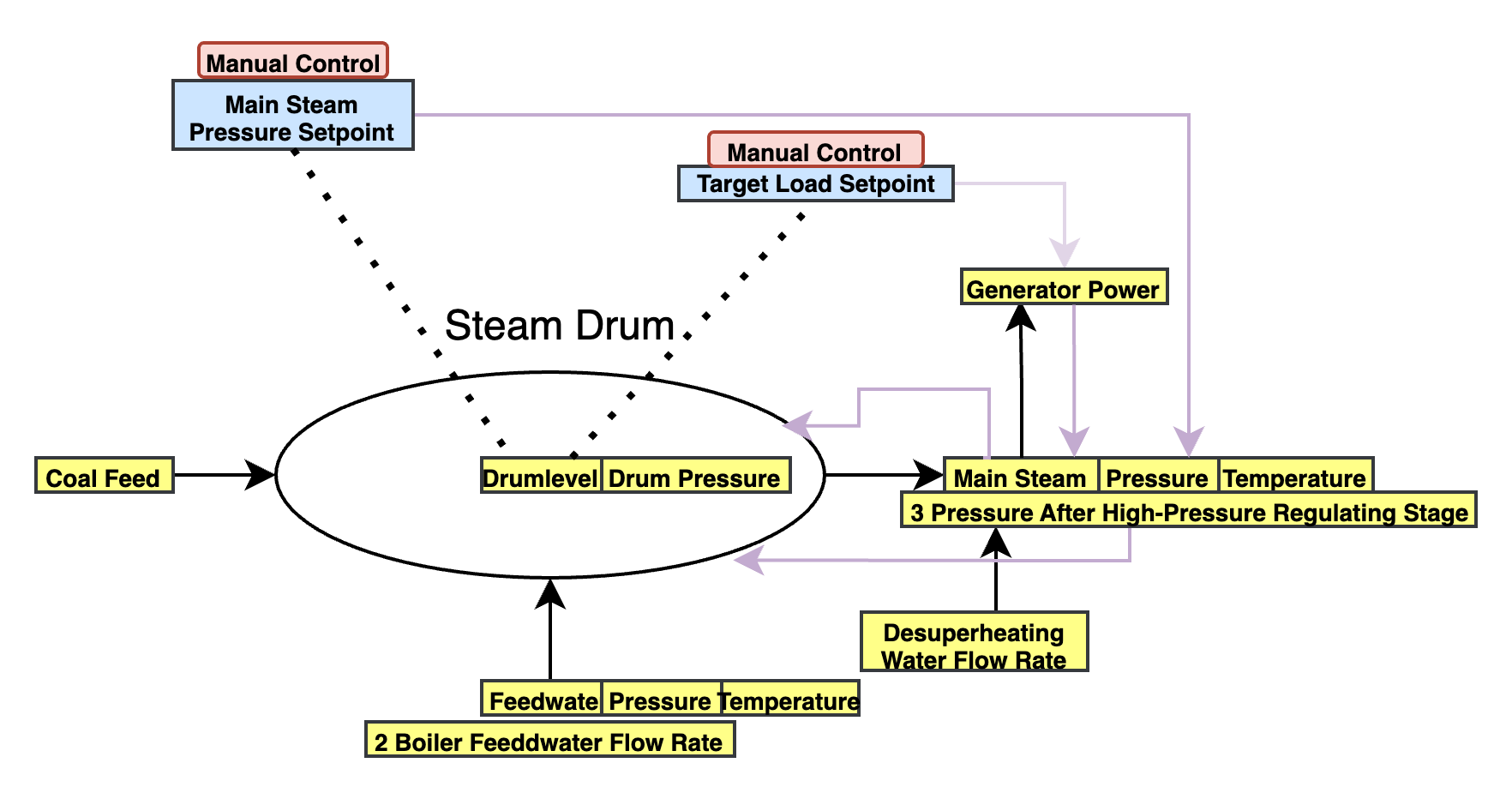}}
\caption{The factors that have a direct causal link to drum level.}
\label{fig:direct_factors}
\vspace{-10pt}
\end{figure}

\subsection{Optimal Time Delay Selection}
To determine the optimal time delay for each variable, we employed a method based on maximum covariance coefficients. 

\begin{equation}
    t^*_{\text{delay}} = \arg \max_{t_{\text{delay}}} \textit{ covariance} (\mathbf{x}_{t - t_{\text{delay}}}, \mathbf{y}_{t})
\end{equation}
where $y$ is the drum level, and $x$ is the factors whose delay needs to be decided. 

Some variables, such as the difference between feedwater pressure and drum pressure, exhibit causal relationships with changes in drum water level but require careful consideration of the delay in their effects. By analyzing the periodic patterns during stable periods, we determined the optimal time delays by aligning peaks and troughs to assess their impact on the subsequent drum water level.

\begin{figure}[htbp]
\centering
\begin{minipage}[t]{\linewidth}
\centering
\includegraphics[width=\linewidth]{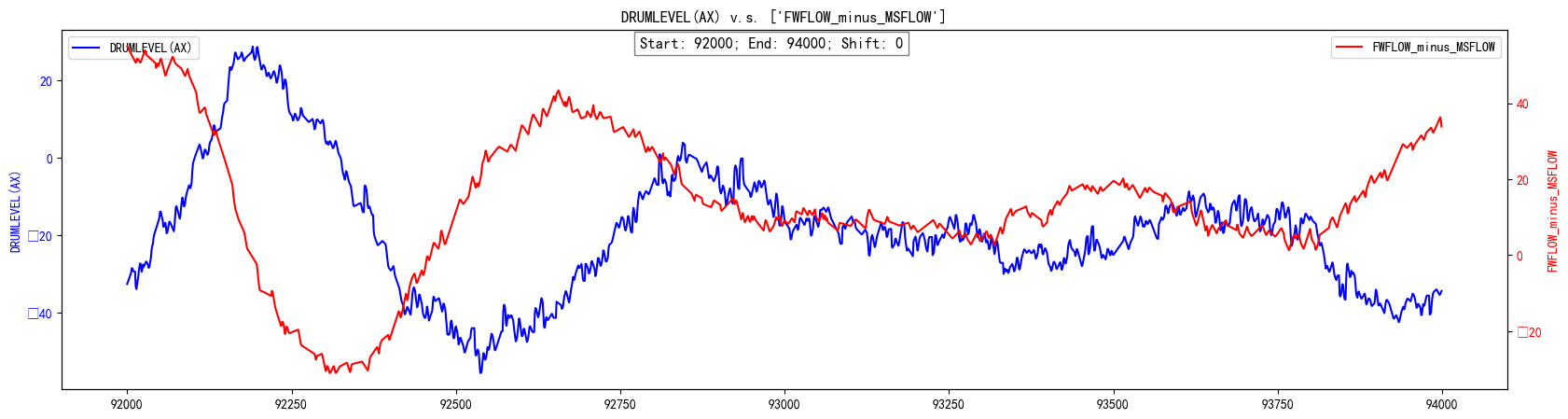}
\end{minipage}

\vspace{20pt} 

\begin{minipage}[t]{\linewidth}
\centering
\includegraphics[width=\linewidth]{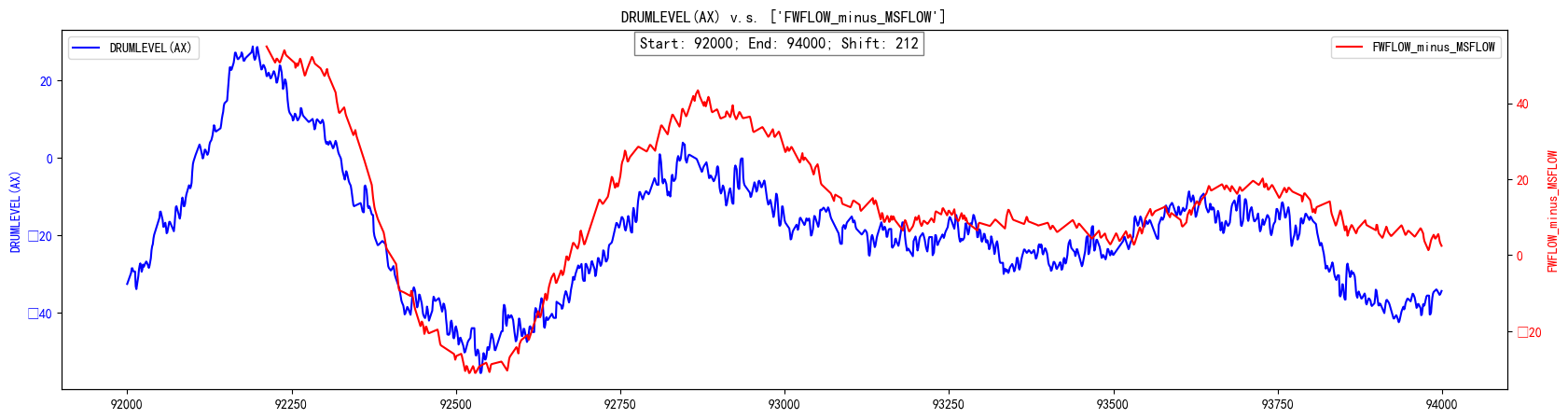}
\caption{Delay Analysis: drum level is in blue, and the factor whose delay is to be calculated, e.g., feedwater flow, is in red. \textbf{Top:} before the shifting, the feedwater flow (red) arrives the peak earlier than drum level (blue); \textbf{Bottom:} After calculating the shifting according to Eq. (2), the optimal delay is 212 seconds and the two lines are synced (meaning their covariance is maximized).}
\end{minipage}
\label{fig}
\vspace{-10pt}
\end{figure}

\subsection{Variable Augmentation}
As mentioned before, after inferring the delay for each factor, a few augmentation schemes will be conducted: both the original variable and the one shifted forward by $t_{\text{delay}}$ will be kept for prediction: namely, both $\mathbf{x}_{t}$ and $\mathbf{x}_{t-t_{\text{delay}}}$ will be used to predict $\mathbf{y}_{t+1}$ (in the case of one step prediction).

\subsection{Model Architecture Design}

For the concatenation of features, we devised a model architecture that accommodates the diverse range of input variables. The Transformer-based model is designed based on a Multi-Input Single-Output (MISO) nonlinear model of the pilot plant. The output will be the prediction of drum water level for the upcoming future time steps. The Transformer model was chosen for its ability to handle sequential data and capture long-range dependencies effectively. We concatenated the engineered features and inputted them into the Transformer layers, allowing the model to learn intricate relationships and patterns in the data. 

\begin{figure}[htbp]
\centerline{\includegraphics[width=\linewidth]{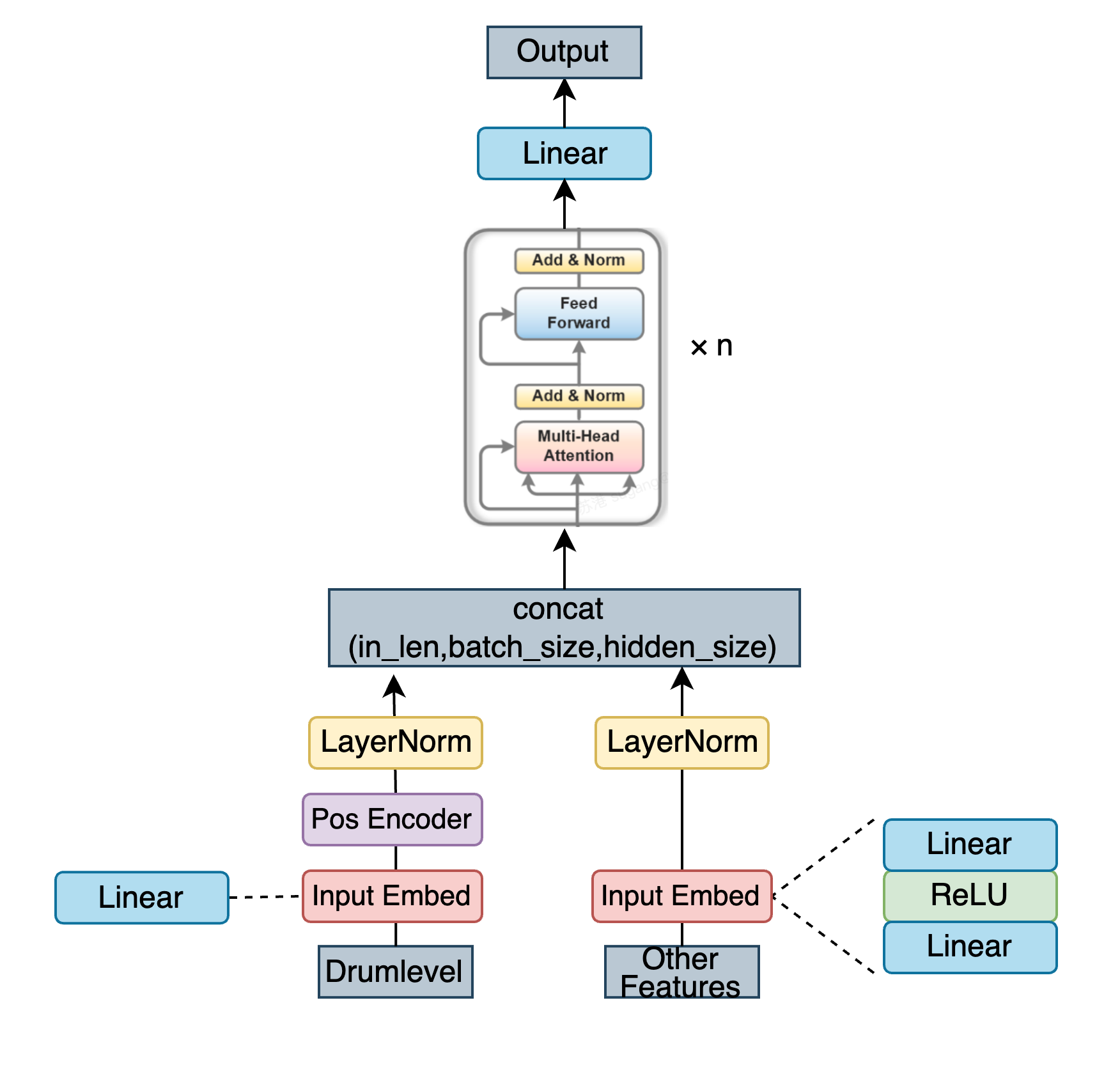}}
\vspace{-10pt}
\caption{Model Architecture.}
\label{fig}
\end{figure}

Let $X = \{x_1, x_2, ..., x_n\}$ represent the input sequence of historical \re{variables}
, where $x_i$ denotes \re{variables}
at time step $i$. Similarly, let $Y = \{y_1, y_2, ..., y_m\}$ denote the target sequence of predicted water level values.

The prediction task can be formulated as follows:

\begin{enumerate}
    \item \textbf{Input Embedding:} The input sequence $X$ is embedded into a continuous representation using an embedding matrix $E$, yielding $\mathbf{X} = \{ \mathbf{x}_1, \mathbf{x}_2, ..., \mathbf{x}_n \}$, where $\mathbf{x}_i$ represents the embedded representation of $x_i$.
    
    \begin{equation}
    \mathbf{x}_i = E(x_i)
    \end{equation}
    
    \item \textbf{Positional Encoding:} Positional encodings are added to the embedded input sequence to provide information about the positions of tokens in the sequence.
    
    \begin{equation}
    \mathbf{x}_i = \mathbf{x}_i + PE(i)
    \end{equation}
    
    \item \textbf{Self-Attention Mechanism:} The Transformer model attends to all input tokens to generate context-aware representations. The attention scores $\mathbf{A}$ are computed using queries $Q$, keys $K$, and values $V$ matrices.
    
    \begin{equation}
    \mathbf{A} = \text{softmax} \left( \frac{QK^T}{\sqrt{d_k}} \right) V
    \end{equation}
    
    \item \textbf{Feedforward Network:} The context-aware representations are passed through feedforward neural networks to capture complex dependencies and generate output representations.
    
    \begin{equation}
    \mathbf{H} = \text{ReLU}(\mathbf{A}W_1 + b_1)
    \end{equation}
    \begin{equation}
    \mathbf{O} = \mathbf{H}W_2 + b_2
    \end{equation}
    
    \item \textbf{Output Prediction:} Finally, the output representations $\mathbf{O}$ are linearly transformed to predict the future steam drum water level values.
    
    \begin{equation}
    \hat{Y} = \mathbf{O}W_o + b_o
    \end{equation}
    
\end{enumerate}

Here, $W_1, W_2, W_o$ represent weight matrices, and $b_1, b_2, b_o$ are bias vectors. $PE(i)$ denotes the positional encoding for token $i$, and $d_k$ represents the dimensionality of the keys and queries.

\subsection{Model Training and Evaluation}\label{SCM}
 The model was trained using historical data with appropriate validation and testing sets. Training involved optimizing model parameters to minimize a specified loss function, such as mean squared error. The trained model was evaluated using various performance metrics, including mean squared error (MSE) and mean absolute error (MAE) and mean absolute percentage error (MAPE), to assess its predictive accuracy and robustness.



\section{Experiments}

\subsection{Data}\label{AA}
For the task of forecasting drum water level in the future next seconds, we gathered a comprehensive dataset comprising 63 variables pertinent to steam drum water level, supplemented by 9 key variables encompassing feedwater, main steam, coal consumption, generator power, and desuperheating water. The dataset covers a time range from January 2nd, 2017, 11:00:01, to January 14th, 2017, 14:16:14, comprising a total of 1,048,574 data points. Data collection involved real-time monitoring and historical records to ensure comprehensive coverage of various operating states and conditions. The visualization presented in Fig. \ref{fig:variables} offers valuable insights into the behavior of all collected variables. 

\begin{table}[t]
\caption{Single Step Prediction Comparison}
\begin{center}
\begin{tabular}{|c|c|c|c|}
\hline
\textbf{ }&\multicolumn{3}{|c|}{\textbf{Performance Evaluation}} \\
\cline{2-4} 
\textbf{Prediction Steps} & \textbf{\textit{MAE}}& \textbf{\textit{MSE}}& \textbf{\textit{MAPE (\%)}} \\

\hline
\textbf{Baseline} & 0.71 & 0.91 & 11.5 \\

\hline
\textbf{LSTM (- delay)} & 0.6811 & 0.8425 & 12.1 \\

\hline
\textbf{LSTM (+ delay)} & 0.6783 & 0.8361 & 11.2 \\

\hline
\textbf{Transformer (- delay)} & 0.6857 & 0.8434 & 12.0 \\

\hline
\textbf{Transformer (+ delay)} & 0.6710 & 0.84 & 11.0 \\

\hline
\end{tabular}
\label{tab1}
\end{center}
\end{table}

\begin{table}[t]
\caption{Comparison of Multiple Prediction Steps}
\begin{center}
\begin{tabular}{|c|c|c|c|}
\hline
\textbf{ }&\multicolumn{3}{|c|}{\textbf{Performance Evaluation}} \\
\cline{2-4} 
\textbf{Prediction Steps} & \textbf{\textit{MAE}}& \textbf{\textit{MSE}}& \textbf{\textit{MAPE (\%)}} \\

\hline
\textbf{Baseline} & & & \\
1 & 0.71 & 0.91 & 12 \\
5 & 2.14 & 7.28 & 42 \\
10 & 2.41 & 9.19 & 50 \\
20 & 3.05 & 14.66 & 61 \\
30 & 3.15 & 15.61 & 66 \\
40 & 3.22 & 16.27 & 69 \\
60 & 3.34 & 17.41 & 79 \\

\hline
\textbf{Transformer} & & & \\
1 & 0.67 & 0.84 & 11 \\
5 & 1.96 & 6.04 & 51 \\
10 & 2.27 & 8.07 & 63 \\
20 & 2.50 & 9.87 & 73 \\
30 & 2.50 & 9.87 & 75 \\
40 & 2.61 & 10.81 & 79 \\
60 & 2.63 & 11.01 & 86 \\
\hline
\end{tabular}
\label{tab1}
\end{center}
\end{table}

\subsection{Benchmark Methods}

Similarly, in the experimental section, we also explored LSTM (Long Short-Term Memory) models. 
\re{Besides, we also employ a history value as a baseline: for example, in one-step prediction, the baseline directly uses $y_t$ from the last time step as the prediction for $y_{t+1}$.}

\begin{figure}[t]
\centerline{\includegraphics[width=\linewidth]{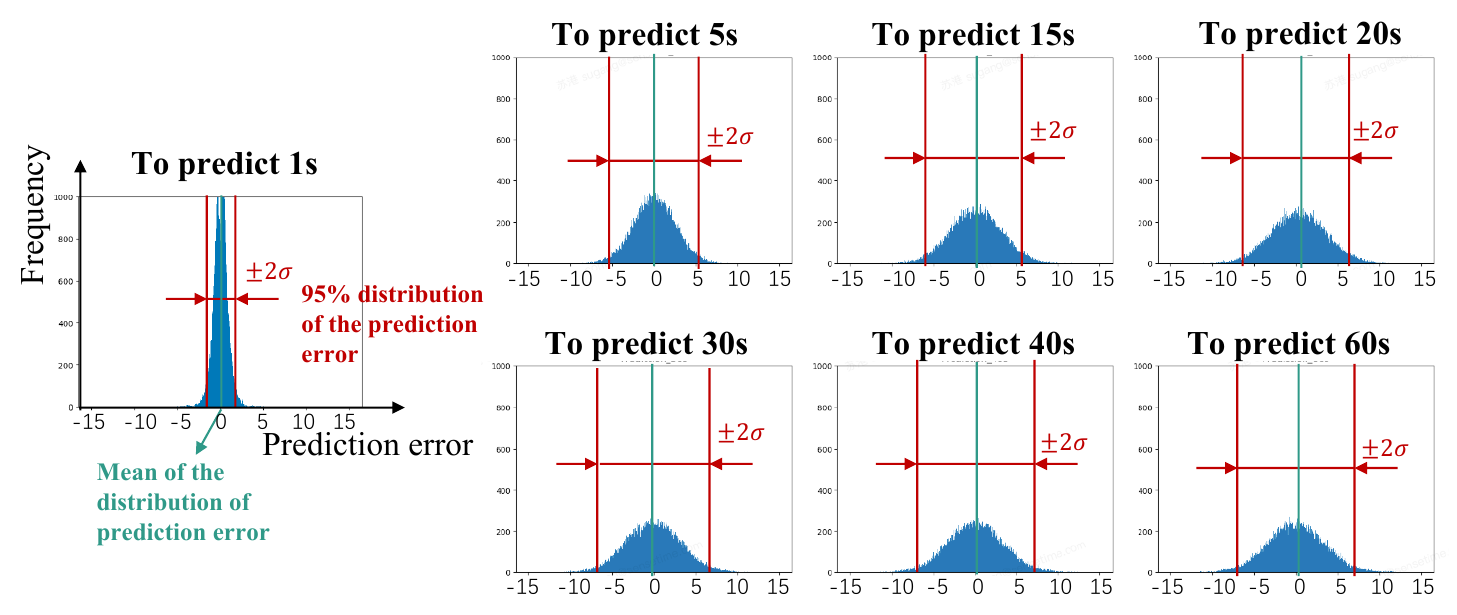}}
\caption{The distribution of prediction error under different prediction horizons.}
\label{fig:error_distribution}
\vspace{-10pt}
\end{figure}

\begin{figure}[t]
\centerline{\includegraphics[width=\linewidth]{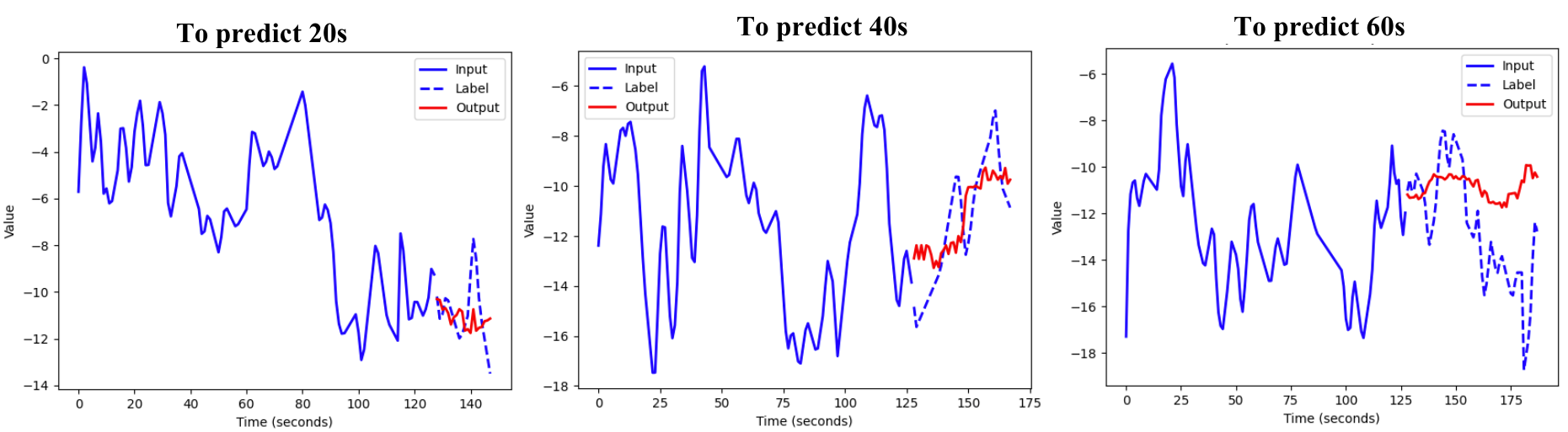}}
\caption{The predicted drum level of the future 20s, 40s, and 60s.}
\label{fig:pred_vis}
\vspace{-10pt}
\end{figure}

\subsection{Results}

As shown in Table I and II, when predicting short-term variations, the performance of Transformer models is comparable to baseline and LSTM. However, as the prediction horizon extends to the long term, the advantages of Transformer architectures gradually become more evident. 
This is because Transformer models leverage self-attention mechanisms, allowing them to attend to relevant parts of the input sequence regardless of distance, thus facilitating better long-term memory encoding \cite{zhou2021informer}. Therefore, the superior performance of Transformer models in long-term prediction tasks can be attributed to their inherent ability to capture and retain long-range dependencies effectively.

\textbf{The Effectiveness of Considering Delay}: As shown in Table 1, we also compare LSTM and Transformer in the setting of considering causal and delay or not; it shows that our design of causal and delay constantly improves the prediction.

\re{
\textbf{Distribution of the Prediction Error}: In Fig. \ref{fig:error_distribution}, we also visualize the distributions of prediciton errors under different prediction horizons. We could observed that our prediciton has been constantly accurate, with mean of the error around 0; moreover, our prediction is also very stable, with the $\pm 2 \sigma$ around $\pm 5$mm. Yet, with longer prediction horizon, the stability sacrifices a bit.

\textbf{Visualization of Predicted Drum Level}: In Fig. \ref{fig:pred_vis}, we can observe that the predicted drum level follows a trend and value that quite respects the ground truth.

\section{Conclusion}
In the paper, we proposed a solid framework to offer an accurate prediction for drum level. The prediction task is rather challenging due to the complex interrelation of the multivariate and their delayed impact on the drum level. Our key design of causal relation analysis, delay inference, and variable augmentation, equipped with a Transformer as the prediction backbone, offers an accurate (mean is around 0) and stable (two sigma range is around 5mm) prediction.
}



\bibliographystyle{plain}
\bibliography{main}

\end{document}